\documentclass[letterpaper, 10 pt, conference]{ieeeconf}
\IEEEoverridecommandlockouts
\usepackage{cite}
\usepackage{amsmath,amssymb,amsfonts}
\usepackage{graphicx}
\usepackage{textcomp}
\usepackage[dvipsnames]{xcolor}
\def\BibTeX{{\rm B\kern-.05em{\sc i\kern-.025em b}\kern-.08em
    T\kern-.1667em\lower.7ex\hbox{E}\kern-.125emX}}
    \usepackage{stackengine}
\usepackage[utf8]{inputenc} 
\usepackage[T1]{fontenc} 
\usepackage{graphicx}
\usepackage{float} 
\usepackage{caption}
\usepackage{url}
\usepackage{mathtools} 
\usepackage{mathrsfs}
\usepackage{colortbl} 
\usepackage{siunitx} 
\usepackage{fancyhdr}
\usepackage[makeroom]{cancel}

\let\labelindent\relax
\usepackage{amsthm}
\usepackage{subcaption}
\usepackage{tikz}
\usetikzlibrary{patterns,angles,quotes,arrows}
\usepackage{cite, enumitem}
\usepackage{algpseudocode}
\usepackage[normalem]{ulem}
\usepackage{booktabs}
\usepackage[belowskip=-15pt,aboveskip=0pt]{caption}

\usepackage{algorithm2e}
\RestyleAlgo{ruled}
\SetKwComment{Comment}{/* }{ */}

\DeclareMathOperator*{\argmin}{arg\,min}

\begin{document}

\title{\Large \textbf{Neural Network-based Fault Detection and Identification for Quadrotors using Dynamic Symmetry}
\thanks{KG and CF are with the Department of Aeronautics and Astronautics at MIT, \texttt{\{kgarg, chuchu\}@mit.edu}.}
}
\author{Kunal Garg \and Chuchu Fan}

\maketitle

\begin{abstract}
Autonomous robotic systems, such as quadrotors, are susceptible to actuator faults, and for the safe operation of such systems, timely detection and isolation of these faults is essential. Neural networks can be used for verification of actuator performance via online actuator fault detection with high accuracy. In this paper, we develop a novel model-free fault detection and isolation (FDI) framework for quadrotor systems using long-short-term memory (LSTM) neural network architecture. The proposed framework only uses system output data and the commanded control input and requires no knowledge of the system model. Utilizing the symmetry in quadrotor dynamics, we train the FDI for fault in just one of the motors (e.g., motor $\# 2$), and the trained FDI can predict faults in any of the motors. This reduction in search space enables us to design an FDI for partial fault as well as complete fault scenarios. Numerical experiments illustrate that the proposed NN-FDI correctly verifies the actuator performance and identifies partial as well as complete faults with over $90\%$ prediction accuracy. We also illustrate that model-free NN-FDI performs at par with model-based FDI, and is robust to model uncertainties as well as distribution shifts in input data.
\end{abstract}


\section{Introduction}



Safety-critical systems are those where violation of safety constraints could result in loss of lives, significant property damage, or damage to the environment. In real-life applications, many cyber-physical control systems are safety-critical, including autonomous cars, unmanned aerial vehicles (UAVs), and aircraft, where safety pertains to keeping the autonomous agent in a predefined safe set away from obstacles and other agents in its environment. In this context, safe control requires finding a control policy that keeps the system within the safe region at all times. As autonomous systems become more complex (thus increasing the likelihood of faults~\cite{lin2020human}), it becomes necessary to explicitly consider the possibility of faults in their actuators which can make it difficult (or even impossible in certain cases) to keep the system safe. Many real-world flight incidents have been attributed to actuator failures such as runaway, sticking, and floating~\cite{abbaspour2020survey}. Such failures have been studied in the field of fault-tolerant control (FTC), which has been applied extensively to applications such as aircraft~ \cite{eltrabyly2021fault, sharifi2010fault, freddi2011feedback, wang2020active}, and spacecraft attitude controls~\cite{zhu2021adaptive, smaeilzadeh2019finite}. 

There is a long history of work in control theory dealing with adaptation to system faults and verification of safe controllers. Due to space limits, we only discuss common FTC techniques for actuator faults. Classical methods include robust control~\cite{slotine_li_1991}; more recent works include robust MPC~\cite{rmpc} and quantitative resilience~\cite{JB2022}. In this paper, we focus on safe fault-tolerant control, where the salient issue is in ensuring that the system will avoid entering an unsafe set despite actuator faults such as loss of control authority or unknown input disturbances. FTC methods are, in general, classified into two categories: active and passive. Active FTC uses detection techniques and a supervisory system to detect the fault and modify the control structure accordingly. Passive FTC relies on a robust compensator to reduce the effects of faults. For a more in-depth explanation of the passive and active FTC theory, see~\cite{abbaspour2020survey}. Many FTC approaches have been presented in the literature to accommodate actuator faults. Fuzzy logic control \cite{er2018novel} and data-driven approaches such as \cite{lin2020human} are used for compensating unknown nonlinear dynamics and actuator faults in multi-agent systems. Feedback linearizing control \cite{lanzon2014flight, freddi2011feedback}, model predictive control \cite{eltrabyly2021fault, rmpc}, sliding mode control \cite{sharifi2010fault}, and adaptive sliding mode control \cite{wang2020active} have been implemented on various nonlinear systems under faults such as a quadrotor subject to one or more motor failures. Adaptive control \cite{zhu2021adaptive} and robust adaptive control \cite{smaeilzadeh2019finite} were studied for linear systems under actuator fault, model uncertainty, and external disturbance. Adaptive fuzzy FTC is presented in \cite{yang2020adaptive} for actuator faults in Markov jump systems. However, FTC-based approaches can be conservative without accurate identification of the source of the fault. Thus, it is essential to design a highly reliable fault detection and identification (FDI) framework that can predict a fault with high accuracy. 

There is a plethora of work on FDI; we refer interested readers to the survey articles \cite{hwang2009survey,puchalski2022uav,fourlas2021survey} that discuss various approaches of FDI used in the literature. In particular, the residual-based method has been used very commonly in prior work, where the expected state or output (under the commanded input and a known system model) and the actual state or output of the system are compared for fault detection. The authors in \cite{guzman2019actuator} study FDI for a linear parameter-varying (LPV) system and use an observer-based FDI that uses the residual data. Such \textit{residual} information requires the knowledge of the system model, and thus, is model-dependent. Another example of a model-based approach is \cite{ouadine2020uav} where model-based output-residual are used instead of state-residuals. The work in \cite{avram2017quadrotor} is capable of handling partial faults but not complete faults in an actuator. Most of the work on adaptation-based FDI uses a linearized model for the system \cite{zhong2018robust,cen2014robust}. 

Neural network (NN)-based verification and system monitoring have been successfully used for FDI. The architecture of the considered NN is very important for such verification problems. Fault detection using system trajectory data can be interpreted as anomaly detection in time-series data, and hence, long-short-term memory (LSTM)-based NNs become a natural choice for FDI as they can effectively handle time-series data \cite{ghrib2020hybrid,provotar2019unsupervised,ergen2019unsupervised}. There is some work on using LSTM-based FDI, e.g., \cite{park2021data,bondyra2022acoustic}, but it is limited to a very narrow class of faults. As noted in \cite{alippi2016model}, prior work on neural network-based model-free FDI relies on reconstruction of the model using artificial neural networks (e.g., \cite{bakhtiaridoust2022model}), or generating the residual information using Kalman filtering (see e.g., \cite{thirumarimurugan2016comparison}). The method in \cite{alippi2016model} also estimates a reduced-order model of the system as an intermediate step. The main disadvantage of model-based FDI methods is that their performance can degrade significantly due to model uncertainties or imperfections in the used model for designing the FDI mechanism and the actual system model. 

To overcome this limitation, in this paper, we design a \textit{model-free} FDI mechanism that only uses the output of the system and the commanded input to the system, and does not use the residual information. The paper's contributions are summarized below:
\begin{itemize}
    \item We present a \textit{truly} model-free approach, where we do not require to either learn the system model or create a reduced-order representation of the model. Instead, we use the system output and the commanded input as the features of a neural network, which directly predicts whether there is an actuator fault.
    \item We consider a variety of partial fault scenarios and leverage the symmetry in quadrotor dynamics to reduce the search space for training and design an NN-FDI trained on the failure of just one motor that is capable of predicting fault in any of the quadrotor motors.
    \item We illustrate through numerical experiments that the model-free FDI mechanism performs at par (and even better in some cases than) the model-based mechanisms. 
    \item We also illustrate the robustness of the proposed method against modeling uncertainties and demonstrate through numerical examples that while the performance of the model-based FDI mechanism drops significantly under model uncertainties, the performance of the designed model-free approach remains the same. 
\end{itemize}

\section{Problem formulation}
We start by presenting the quadrotor dynamics. The quadrotor dynamics can be written compactly as: 
\begin{subequations}\label{nominal}
\begin{align}
 \dot x & = f(x) + g(x)u, \\
 y & = \rho(x),
\end{align}
\end{subequations}
for state $x \in \mathcal X$, control input $u \in \mathcal U$, and state and control sets $\mathcal X\subset\mathbb R^n$ and $\mathcal U\subset \mathbb R^m$, respectively. Here, $\rho:\mathbb R^{12}\to\mathbb R^6$ is the output map consisting of the position and the attitude vector of the quadrotor, i.e., $y = (p_x, p_y, p_z, \phi, \theta, \psi)$. Such an output model is realized using a 6DOF Inertial Measurement Unit (IMU) output. The 6-DOF quadrotor dynamics are given in \cite{budaciu2019evaluation} with $x\in \mathbb R^{12}$ consisting of positions, velocities, angular positions and angular velocities, and $u\in \mathbb R^4$ consisting of the thrust at each of four motors
:{
\begin{subequations}
\begin{align}
    \dot p_x = & ~ \big(c(\phi)c(\psi)s(\theta)+s(\phi)s(\psi)\big)w\nonumber\\
    &-\big(s(\psi)c(\phi)-c(\psi)s(\phi)s(\theta)\big)v + u c(\psi)c(\theta) \\
    \dot p_y = & ~ \big(s(\phi)s(\psi)s(\theta)+c(\phi)c(\psi)\big)v\nonumber\\
    &-\big(c(\psi)s(\phi) -s(\psi)c(\phi)s(\theta)\big)w + u s(\psi)c(\theta)\\
    \dot p_z = & ~ w~c(\psi)c(\phi)-u~s(\theta)+v~s(\phi)c(\theta)\\
    \dot u = & ~ r~v-q~w+g~s(\theta)\\
    \dot v = & ~ p~w - r~u -g~s(\phi)c(\theta)\\
    \dot w = & ~ q~u-p~v + \frac{U_1}{m}-g~c(\theta)c(\phi)\\
    \dot \phi = & ~ r\frac{c(\phi)}{c(\theta)} + q\frac{s(\phi)}{c(\theta)} \\
    \dot \theta = & ~ q~c(\phi)-r~s(\phi)\\
    \dot \psi = & ~ p +r~c(\phi)t(\theta) + q~s(\phi)t(\theta) \\
    \dot r = & ~ \frac{1}{I_{zz}}\big(U_2-pq(I_{yy}-I_{xx})\big)\\
    \dot q = & ~ \frac{1}{I_{yy}}\big(U_3-pr(I_{xx}-I_{zz}) \big)\\
    \dot p  = & ~ \frac{1}{I_{xx}}\Big(U_4+qr(I_{zz}-I_{yy}) \Big)
\end{align}
\end{subequations}}\textnormal
where $m, I_{xx}, I_{yy}, I_{zz}, k_r, k_t>0$ are system parameters, $g = 9.8$ is the gravitational acceleration, $c(\cdot), s(\cdot), t(\cdot)$ denote $\cos(\cdot), \sin(\cdot), \tan(\cdot)$, respectively,  $(p_x, p_y, p_z)$ denote the position of the quadrotor, $(\phi, \theta, \psi)$ its Euler angles and $u = (U_1, U_2, U_3, U_4)$ the input vector consisting of thrust $U_1$ and moments $U_2, U_3, U_4$.

The relation between the vector $u$ and the individual motor speeds is given as {\small 
\begin{align}
    \begin{bmatrix}U_1\\ U_2 \\ U_3\\ U_4 \end{bmatrix}\!=\! \begin{bmatrix}C_T & C_T& C_T& C_T\\
    -dC_T\sqrt{2} & -dC_T\sqrt{2} & dC_T\sqrt{2} & dC_T\sqrt{2}\\
    -dC_T\sqrt{2} & dC_T\sqrt{2} & dC_T\sqrt{2} & -dC_T\sqrt{2}\\
    -C_D & C_D  & -C_D & C_D
    \end{bmatrix} \!\begin{bmatrix}\omega_1^2\\ \omega_2^2\\\omega_3^2\\\omega_4^2\end{bmatrix}\!,
\end{align}}\normalsize
where $\omega_i$ is the angular speed of the $i-$th motor for $i\in \{1, 2, 3, 4\}$, $C_D$ is the drag coefficient and $C_T$ is the thrust coefficient. These parameters are given as: $I_{xx} = I_{yy} = 1.395\times 10^{-5}$ kg-$\textnormal{m}^2$, $I_{zz} = 2,173\times 10^{-5}$ kg-$\textnormal{m}^2$, $m = 0.0299$ kg, $C_T = 3.1582\times 10^{-10}$ N/rpm$^2$, $C_D = 7.9379\times 10^{-12}$ N/rpm$^2$ and $d = 0.03973$ m (see \cite{budaciu2019evaluation}).


In this paper, we consider an actuator fault occurring at some unknown time $t_f \geq 0$:
\begin{align}\label{eq: fault model}
    u(t,x) =\begin{cases} \pi(t,x) & \textnormal{if} \quad t\leq t_{F}; \\
    \textnormal{diag}(\Theta)~\pi(t,x)
    & \textnormal{if}\quad  t>t_F,
    \end{cases},
\end{align}
where $\Theta  = [0, 1]^m\in \mathbb R^m$ is the vector denoting whether an actuator is faulty or not, and $\textnormal{diag}:\mathbb R^m\rightarrow\mathbb R^{m\times m}$ maps a vector in $\mathbb R^m$ to a diagonal matrix in $\mathbb R^{m\times m}$. If the $i-$th actuator is faulty, then $\Theta_i \in [0, 1)$ and the rest of the elements of $\Theta$ are 1. The problem statement is to design an NN-based FDI $\Theta_{NN}$ that correctly predicts and identifies which actuator has a fault and what is the degree of the fault. 

\begin{figure}[b]
    \centering
    \includegraphics[width=0.9\columnwidth]{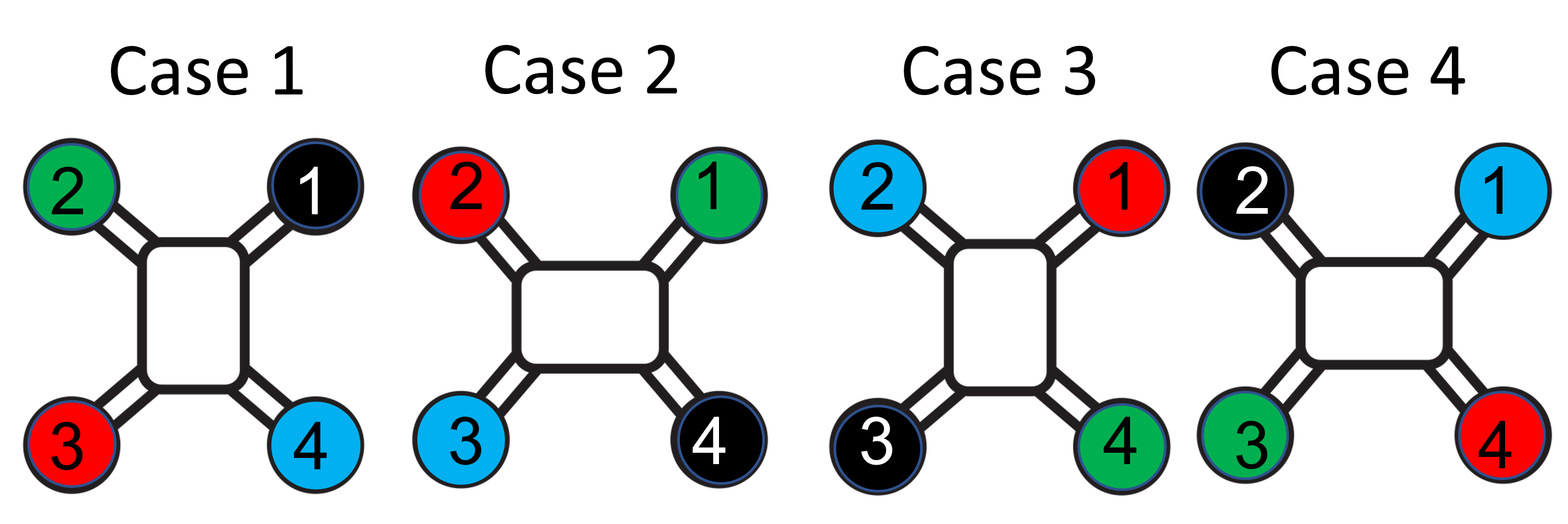}
    \caption{The four cases used in fault-prediction. Each next case is generated through a 90 degrees anti-clockwise axes rotation from the previous model (resulting in a 90-degree clockwise configurational rotation).}
    \label{fig:quad rotate cases}
\end{figure}

\section{Neural Fault-detection and Isolation}\label{sec: NN FDI}
\subsection{Model-free FDI}
The faults must be detected correctly and promptly for the safe recovery of the system. We use a learning-based approach to design a fault-detection mechanism. Let $\Theta \in [0, 1]^m$ denote the fault vector, where $\Theta_i < 1$ indicates that $i-$th actuator is faulty, while $\Theta_i = 1$ denotes it is not faulty. Let $\Theta_{NN}: \mathbb Y\times \mathbb U \rightarrow\mathbb R^m$ be the \textit{predicted} fault vector, parameterized as a neural network. Here, $\mathbb Y = \{x(\cdot) \; |\; y(\cdot) = \rho(x(\cdot))\}$ is a function space consisting of trajectories of the state vector, and $\mathbb U = \{u(\cdot) \; |\; u(\cdot)\in \mathcal U\}$ is a function space consisting of input signals. To generate the residual data, the knowledge of the system model is essential, which makes the residual-based approach model dependent. This is the biggest limitation of this approach, as modeling errors can lead to severe performance issues in fault detection due to model uncertainties. To overcome this, we propose a model-free NN-based FDI mechanism that only uses $(y, u)$ as the feature data, i.e., it does not require the model-based residual information. For a given time length $T>0$, at any given time instant $t \geq T$, the NN function $\Theta_{NN}$ takes a finite trace of the system trajectory $x(t-\tau)|_{\tau = 0}^T$ and the \textit{commanded} input signal $u(t-\tau)_{\tau = 0}^T$ as input, and outputs the vector of predicted faults.

\begin{figure}[t]
    \centering
    \includegraphics[width=\columnwidth]{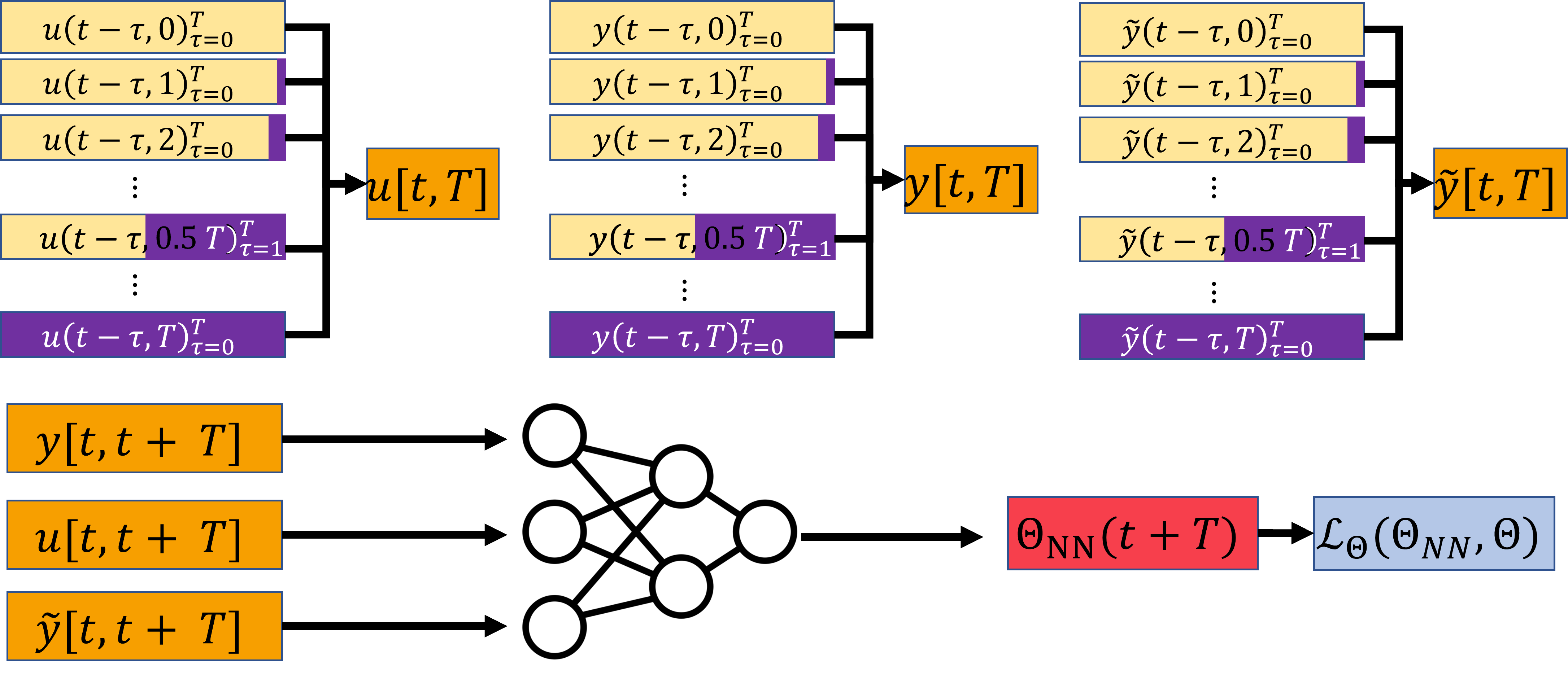}
    \caption{General neural-network architecture for failure prediction. The training data includes all possible trajectories with different lengths of faulty input (the violet color represents the portion of the trajectory with faulty input).}
    \label{fig:NN training data for fault predictor}
\end{figure}

Using the symmetry of the quadrotor (i.e., $I_{xx} = I_{yy}$), it is possible to only learn the fault-detector for one of the faulty actuators and detect which motor is faulty using rotational invariance. Let us define color-coding for the four motors in the original configuration (i.e., case 1): 
\begin{align*}
    &\#1 \rightarrow \textnormal{Black} \quad \#2 \rightarrow \textnormal{Green} \quad \#3 \rightarrow \textnormal{Red} \quad \#4 \rightarrow \textnormal{Blue}
\end{align*}
During the training, without loss of generality, we assume that the green motor is faulty. Now, if instead, another motor is faulty, then a state-transformation map can be defined as $\Phi(n) = \begin{bmatrix}R_\theta & \mathbf{0} & \mathbf{0} & \mathbf{0}\\ \mathbf{0} & R_\theta & \mathbf{0} & \mathbf{0} \\ \mathbf{0} & \mathbf{0} & R_\theta & \mathbf{0}\\ \mathbf{0} & \mathbf{0} & \mathbf{0} & R_\theta\end{bmatrix}$, with $\theta = \frac{\pi}{2}, \pi, \frac{3\pi}{2}$ for n = 3, 4 and 1, respectively, where $R_\theta = \begin{bmatrix}
\cos(\theta) & \sin(\theta) & 0 \\ -\sin(\theta) & \cos(\theta) & 0 \\ 0 & 0 & 1\end{bmatrix}$. Thus, for case \#4, the black motor acts as motor \#2 in the original configuration (see Figure \ref{fig:quad rotate cases}). As a result, if the black motor is faulty, and the fault-predictor is trained to detect that motor \#2 is faulty (i.e., green motor in the original configuration), then case 4 will give the correct prediction (see Figure \ref{fig:prediction accuracy 4 cases}).


\subsection{Model-based FDI}
For model-based FDI mechanisms, the residual data is also required as an additional feature to the NN. The error vector $\tilde y$ (commonly known as residual in the FDI literature) is defined as the stepwise error between the actual state of the system with potentially faulty actuators and the state of the system assuming no faults, i.e., $\tilde y(t) = y(t) - \bar y(t)$ where $y$ is the output of the actual system \eqref{nominal} with faulty input and $\bar y$ is the output of reference model without fault $\dot {\bar x} = f(\bar x) + g(\bar x)u, \bar y(t) = \rho(\bar x(t))$, with $\bar y(k\tau) = y(k\tau))$, $k = 0, 1, 2, \dots$, where $\tau >0$ is sampling period for data collection. In most of the prior literature, either $\tilde x$ (in state-based methods) or $\tilde y$ (in output-based methods) is used for designing FDI. However, that requires the availability of the model for computing the residuals, which makes such approaches very restrictive. 

\subsection{Training data}
For training, the trajectory data is collected where actuator $\#2$ is partially faulty with $\theta_2 \in \{0, 0.1, 0.2, \cdots, 0.9, 1\}$. Let $d = 11$ denote the number of faulty scenarios for motor $\#2$. At each time instant $t\geq T$, it is possible that only a portion of the trajectory is generated under a faulty actuator. That is, the possible input to the system is $u(t-\tau, T_f)_{\tau = 0}^T \coloneqq [u(t-T), u(t-T + 1), \cdots, u_f(t-T + T_f), u_f(t-T + T_f + 1), \cdots, u(t)]$ (with the corresponding system output $y(t-\tau, T_f)_{\tau = 0}^T$ and the residual $\tilde y(t-\tau, T_f)_{\tau = 0}^T$), where $T_f\in [0, T]$ dictates the time instant when the fault occurs. Thus, the NN for fault prediction must be trained on all possible combinations of occurrences of fault. Hence, our training data includes $\bigcup\limits_{T_f\in 0}^T\big(y(t-\tau, T_f)_{\tau = 0}^T, \tilde y(t-\tau, T_f)_{\tau = 0}^T, u(t-\tau, T_f)_{\tau = 0}^T\big)$ (see Figure \ref{fig:NN training data for fault predictor}). In every training iteration, we generate $N_{traj} = N_1\times d \times T_f$ trajectories, so that we have $N_1 > 0$ trajectories for faults in each of the $d$ faults in motor $\#2$ with all possible lengths of trajectories under one faulty actuator in $[0, T_f]$. In particular, we consider discrete fault values $\Theta_2\in \{0, 0.1, 0.2, \cdots, 0.9\}$ and thus, $d = 11$ with 10 values for faults and 1 for non-faulty cases. The training data is generated by randomly choosing initial conditions $\{x(0)\}_1^{N_1}$ and rolling out $d$ trajectories for each of the sampled conditions under $d$ possible $\Theta_2$ values. This enables the NN to distinguish between various kinds of fault scenarios since it obtains trajectories under all considered fault scenarios with the same initialization. 

\begin{algorithm}[t]
\caption{Learning framework for FDI}\label{alg:learn_algo}
\KwData{$iter_M, N, N_{bs}, N_{buf}, N_M, N_m$ }
\KwResult{$\Theta_{NN}$}
Initialize $\Theta_{NN}$ as NNs; \Comment{LSTM-NN}

$\{y, u, \tilde y\}_{buf} = \emptyset$

\While{$iter \leq iter_M$ or (\textnormal{loss} $<10^{-3}$ and $iter\geq N_m$}{

  Sample $\{x_0\}_1^{N_1}$ 

    \For{$i \in N_1$}{
  Roll out $d$ trajectories under $\Theta_2\in [0, 1]$
  
  $\{y, u, \tilde y\}_{buf}= \{y, u, \tilde y\}_{buf}\cup \{y, u, \tilde y\}_1^{N_1}$

  }

    $\{z\} = \{y, u, \tilde y\}_{buf}$
    
    $\{z\}_{train} = \{z\}[-N_{buf}:]$
    
    \While{$\{z\}_{train} \neq \emptyset$}{
    $\{z\}_{train, bs}$ = Sample $N_{bs}$ trajectories from $\{z\}_{train}$ 
    
    loss = $\mathcal L_{\Theta}^{MF}$ \Comment{$\mathcal L_{\Theta}^{MB}$ for model-based} 
    
    train $\Theta_{NN}$ 
    
    $\{z\}_{train} = \{z\}_{train} \setminus\{z\}_{train, bs}$
    }
}
\end{algorithm}

The loss function for model-free FDI training is defined as \normalsize
{
\begin{align}
    \mathcal L_{\Theta}^{MF} = \frac{1}{N_{traj}} \sum_{j = 1}^{N_{taj}}\Big[||\Theta_{j, NN}\big(y_j(\cdot), u_j(\cdot)\big) - \Theta_j\| - \epsilon\Big]_+,
\end{align}}\normalsize
while model-based FDI is given as:
\begin{align}
    \mathcal L_{\Theta}^{MB} = \frac{1}{N_{traj}} \sum_{j = 1}^{N_{taj}}\Big[||\Theta_{j, NN}\big(y_j(\cdot), u_j(\cdot), \tilde y_j(\cdot)\big) - \Theta_j\| - \epsilon\Big]_+,
\end{align}
where $\Theta_j\in\{1\}\times [0, 1] \times\{1\}\times \{1\}$ is the fault vector used for generating the data for the $j-$th trajectory and $0 <\epsilon \ll 1$. In each training epoch, we generate $N = 200 \times 11 \times 100 $ trajectories of length $T_f$ and maintain a buffer of $1.5$~M trajectories. We train the NN until the loss reduces to $10^{-3}$. We use a Linear-Quadratic Regulator (LQR) input to generate the training data. In our experiments, we illustrate that the trained NN is highly robust to the kind of input used for trajectory generation and can predict fault with the same accuracy for the trajectories generated by Control Barrier Functions (CBF)-based quadratic programs (QPs) which are commonly used for maintaining safety \cite{Ames2014}. During training, we optimize the loss function using stochastic gradient descent, and we train the pre- and post-fault networks separately. The number of trajectories in the buffer is capped at $N_{buf}$ so that once the maximum number of trajectories are collected, the earlier trajectories are dropped from the buffer. The training is performed either till the number of iterations reaches $N_M > 0$, or the loss drops below $10^{-3}$ after at least $N_m < N_M$ training epochs. 
During each training epoch, we use a batch size of 50000 trajectories and perform 500 iterations of training on all the buffer data. The learning algorithm is summarized in Algorithm \ref{alg:learn_algo}

\begin{figure}[b]
    \centering
    \includegraphics[width=\columnwidth]{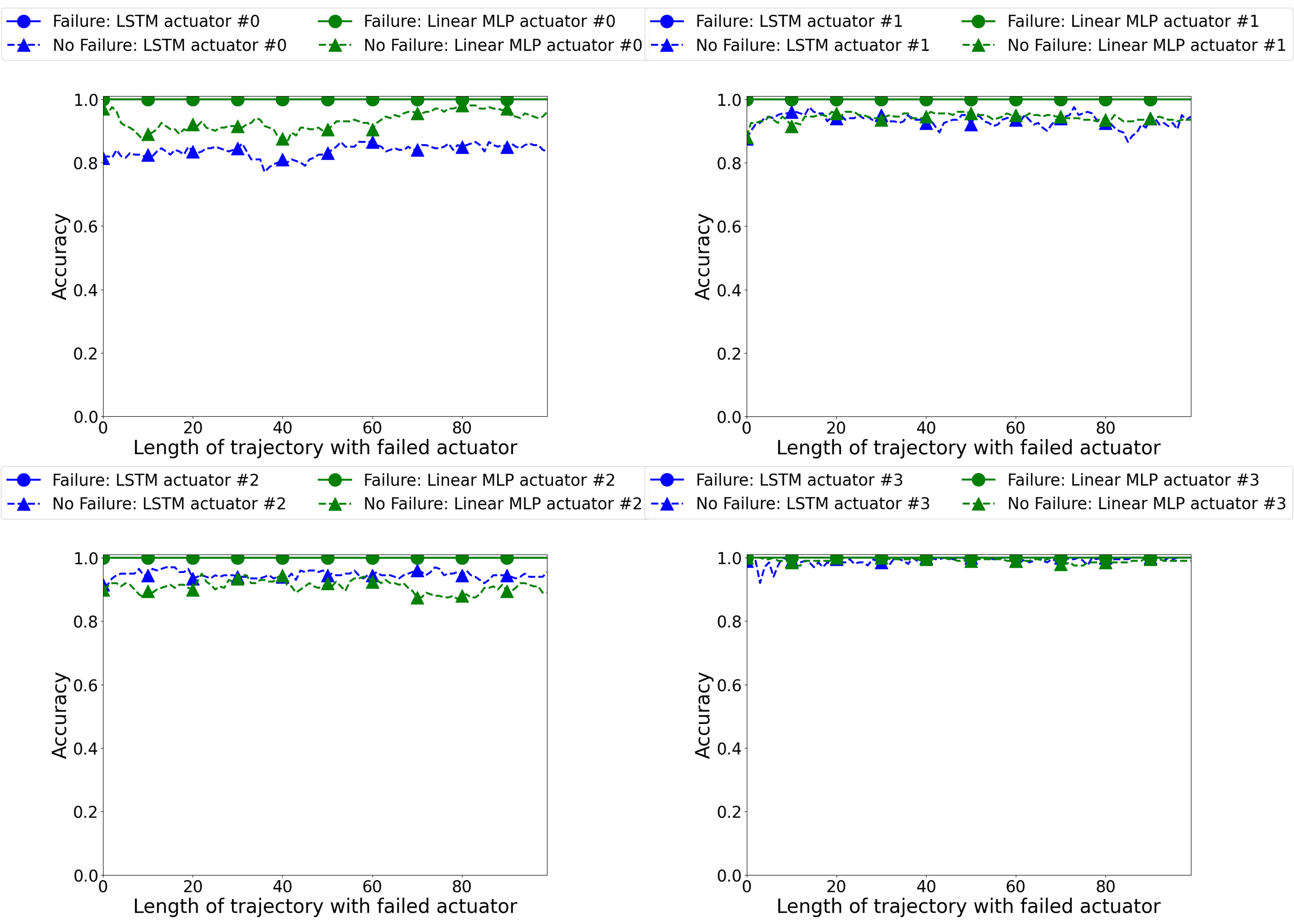}
    \caption{Failure prediction accuracy for faults in different motors: \textbf{Top-left}: Failure in motor \#1, \textbf{Top-right}: Failure in motor \#2, \textbf{Bottom-right}: Failure in motor \#3 and \textbf{Bottom-left}: Failure in motor \#4. This illustrates that with one trained FDI, it is possible to predict failure in any of the actuators with high prediction accuracy. }
    \label{fig:prediction accuracy 4 cases}
\end{figure}

\section{Numerical evaluations} 
The primary objective of our numerical experiments is to evaluate the effectiveness of our method in terms of fault detection. We consider an experimental case study involving the Crazyflie quadrotor with a fault in motor $\#2$. First, we evaluate the correctness of the fault prediction by the 4 cases explained in Figure \ref{fig:quad rotate cases}. A fault is predicted if $\min\limits_i\min\limits_{n}\Theta_{i, NN}(\Phi_n(x)) < \Theta_{tol}$, where $\Theta_{tol} = 0.2$. In this case, we only consider the case when $\Theta_i =0$. The predicted faulty actuator is given by the $\argmin$ of $\Theta_{NN}(\Phi_{n^*}(x))$, where $n^*$ is the rotation index for which $\Theta_{i, NN}$ is below the tolerance. The experiments are run to check the prediction accuracy of the NN-based FDI mechanism for various lengths of data with failed actuators between 0 and $T_f = 100$. We compute the prediction accuracy when there is a fault as well as when there is no fault. We sample 10000 initial conditions randomly from the safe set $\mathcal X_{safe}$ to generate trajectories for test data, where 2000 trajectories are generated for each of the faults and 2000 trajectories are generated without any fault. Each trajectory is generated for 200 epochs with fault occurring at $t = 100$. We feed the moving trajectory data $(x(k-100,k), u(k-100,k), \tilde x(k-100,k)$ to the trained NN-based FDI starting from $k = 100$. For a given $k\in [100, 200]$, the portion of trajectory data with faulty actuator is $k-100$. Figure \ref{fig:prediction accuracy 4 cases} illustrates that the prediction accuracy for correct fault prediction for each of the cases is above 80\%. This illustrates that it is possible to effectively predict faults in all motors with an FDI trained with faults in just one of the motors.

\begin{figure}[t]
    \centering
    \includegraphics[width=\columnwidth]{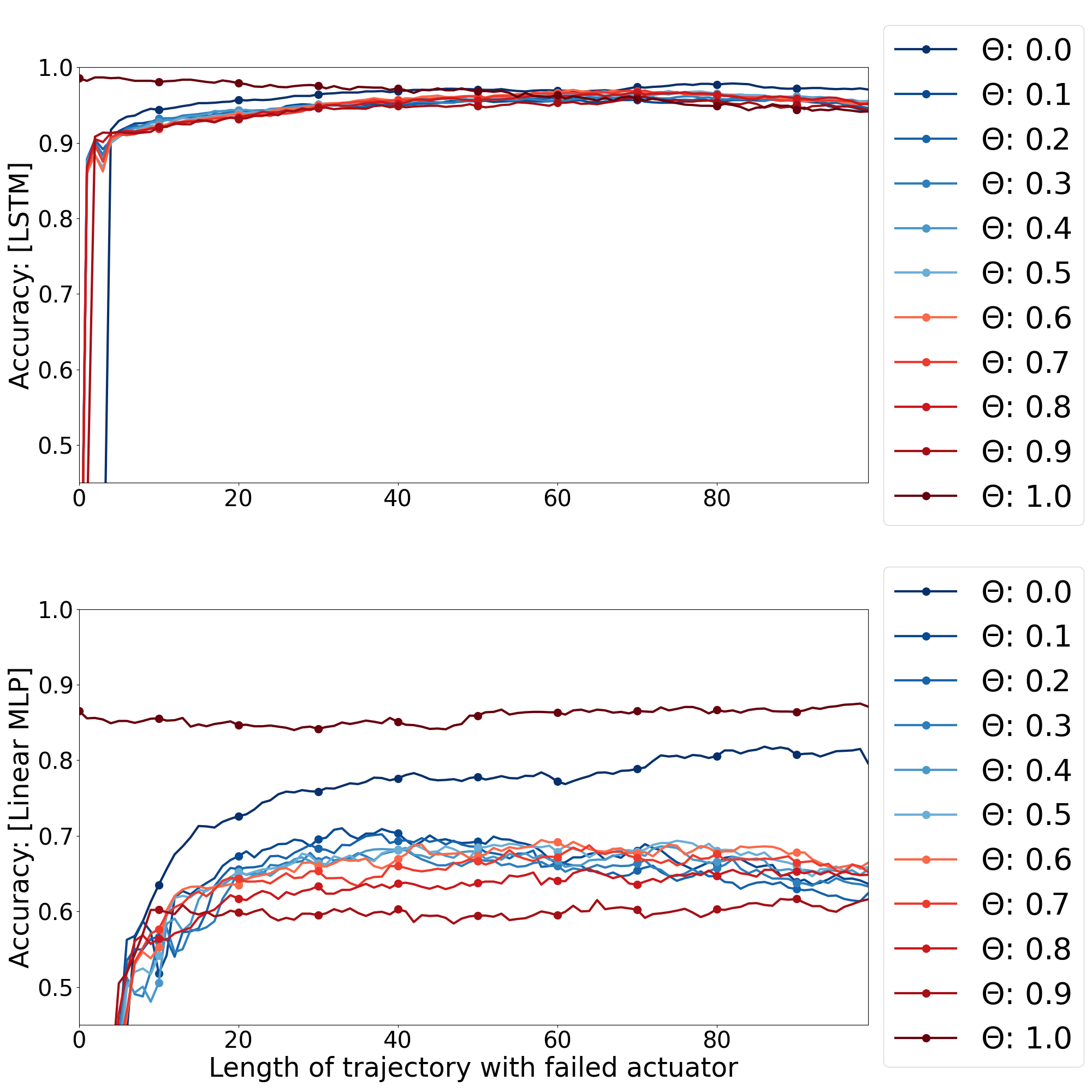}
    \caption{Comparison of prediction accuracy of two NN architectures: a multi-layer perceptron (MLP) with linear layers
    and a long-short-term memory (LSTM) NN for various fault parameters $\Theta\in [0, 1]$. }
    \label{fig:comparison NN}
\end{figure}

Next, we evaluate prediction accuracy for different fault values $\Theta_2\in [0, 1]$. For this experiment, we compare the performance of two different types of NN architectures for model-free FDIs, namely, a multi-layer perceptron (MLP) with 1 input layer, 4 hidden layers, and 1 output layer
and a long-short-term memory (LSTM) where the LSTM layer is followed by 2 linear layers. In each of the NN architectures, we use $N\times$128 as the size of the input layer with $N$ being the size of the features, hidden layer(s) of size 128$\times 128$ followed by a hidden layer of size 128$\times$64 and an output layer of size 64$\times m$. Note that $N = (2p + m)\times T_f$ for the FDI with all the data, $p \times T_f$ for the FDI with just the residual data, and $N = (p + m) \times T_f$ for the model-free FDI mechanism. Figure \ref{fig:comparison NN} plots the prediction accuracy for various values of $\Theta_2$ for the two considered NN architectures. It can also be observed that LSTM-based NN FDI can accurately identify each of the faults while MLP-based FDI has a very low prediction accuracy.

\begin{figure}[t]
    \centering
    \includegraphics[width=\columnwidth]{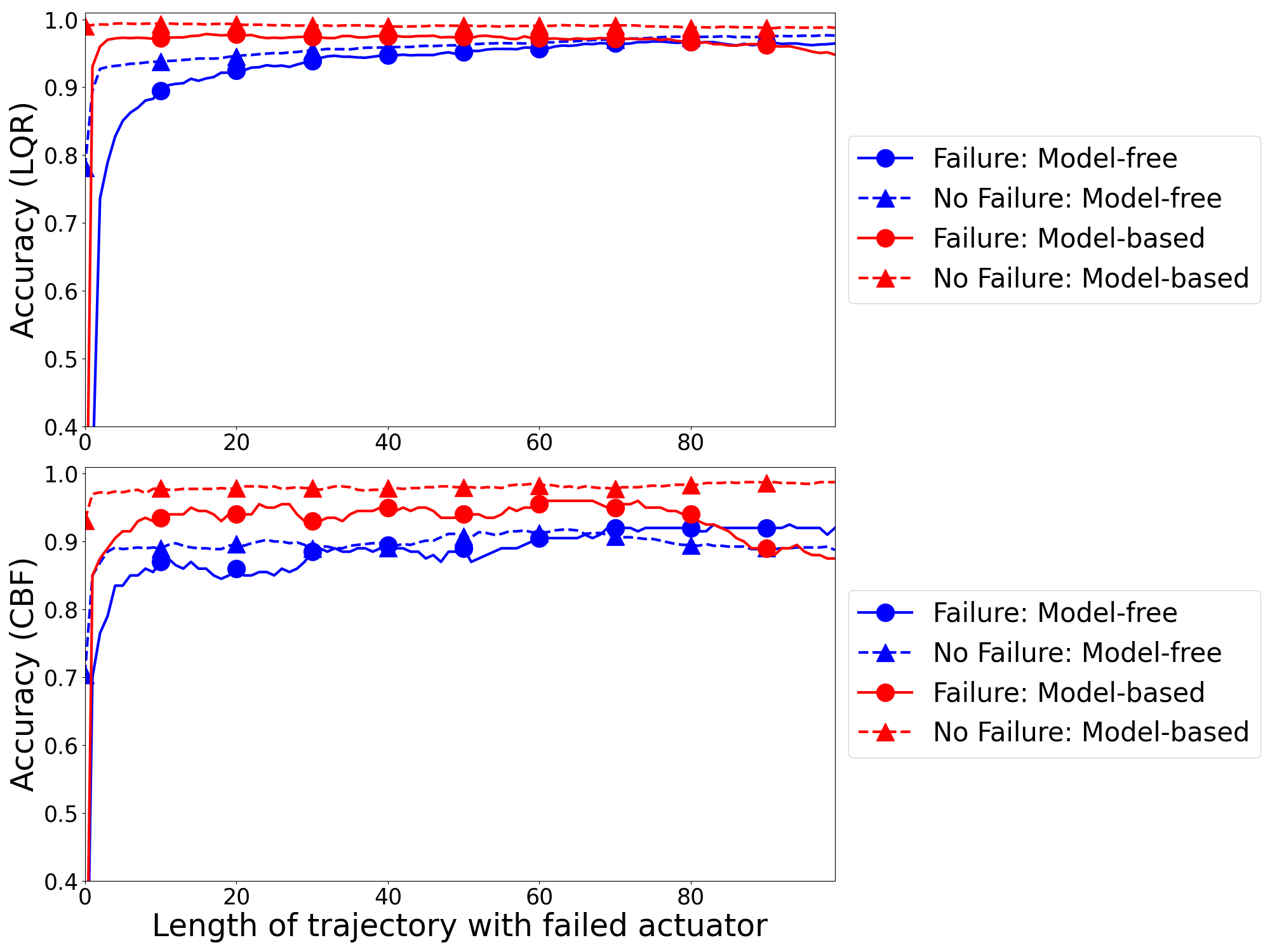}
    \caption{Failure prediction accuracy for CBF-QP input (solid lines) and LQR input (dashed lines). The performance of model-free (Ours) FDI with data $(y, u)$ is shown in blue, while the one with all the data $(y, u, \tilde y)$ is in red.}
    \label{fig:res only vs all}
\end{figure}

We use an LQR input to generate the training data since solving a CBF-based QP is relatively slower for collecting a sufficient amount of training data. In our experiments, we illustrate that the trained NN is highly robust to the kind of input used for trajectory generation and can predict fault with the same accuracy for the trajectories generated by CBF-based QPs. For this experiment, we compare the prediction accuracy of the model-free NN-FDI ($\theta_{NN}(y, u)$) and the model-based FDIs ($\theta_{NN}(y, u, \tilde y)$). Figure \ref{fig:res only vs all} shows the prediction accuracy of the model-based FDIs. It can also be seen that the model-free FDI mechanism can perform at par (even better) than the model-based FDI mechanism with features ($(y, u, \tilde y)$). Based on this observation, we can infer that a model-free FDI mechanism can be used with very high confidence. In this case, it is crucial to note that the trained fault predictor is highly robust with respect to the input data. In particular, to accelerate the learning process, a very simple LQR controller is used, where the nonlinear system dynamics are linearized about the origin, and a constant LQR gain is used. However, as can be seen from Figure \ref{fig:res only vs all}, the prediction generalizes to the CBF-based QP controller just as well and has a similar high prediction accuracy. 

\begin{figure}[t]
    \centering
    \includegraphics[width=1\columnwidth]{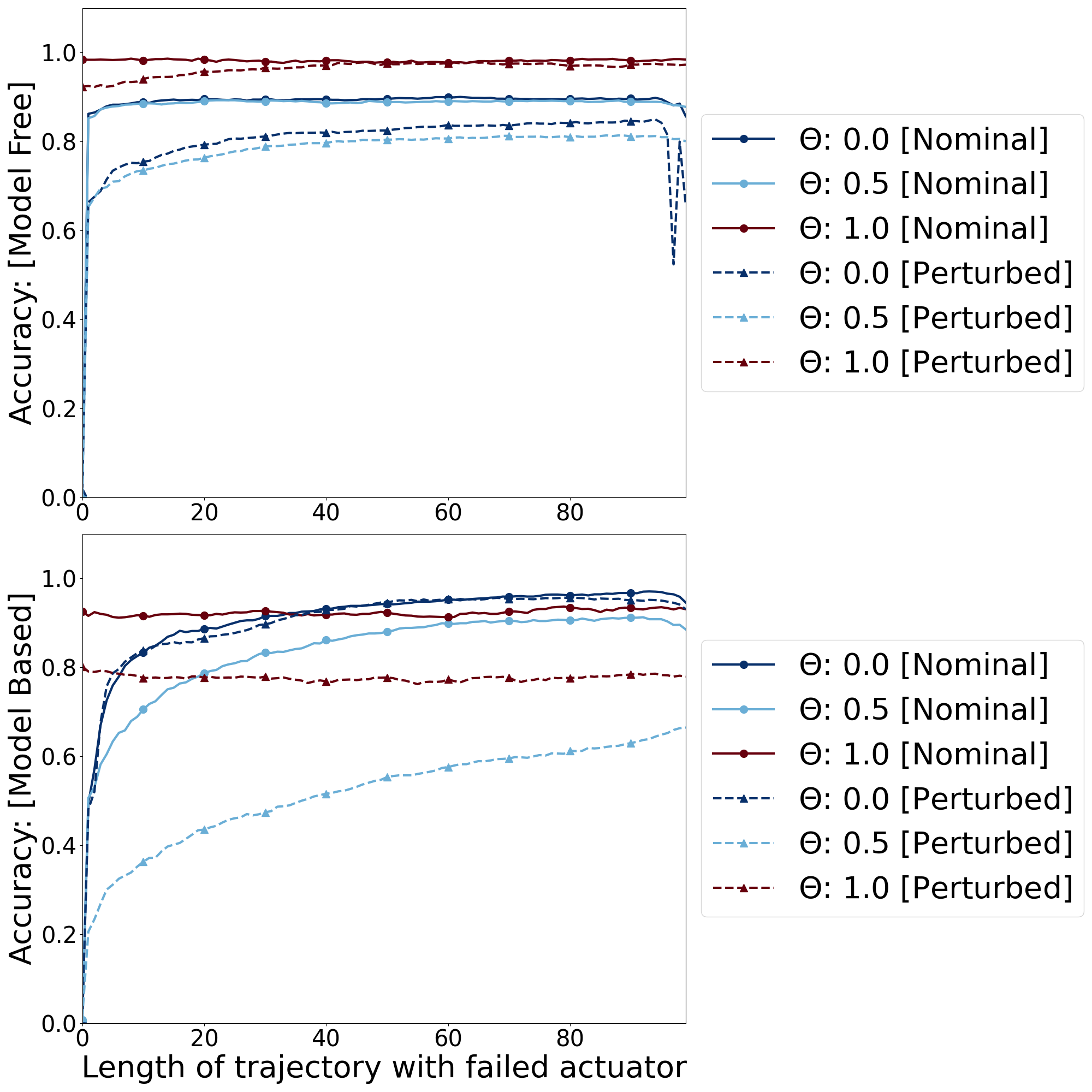}
    \caption{Comparison of prediction accuracy of model-based and model-free FDI mechanisms with perturbation in system parameters.}
    \label{fig:comparison NN pert}
\end{figure}

Finally, we study the effect of change in model parameters (such as the inertia matrix, etc.) on the prediction accuracy of the FDI mechanisms. For this experiment, we changed the system parameters by more than 40\% (see Table \ref{tab:param table}).
\begin{table}[b]
    \centering
    \caption{Nominal and perturbed quadrotor parameters.}
    \hspace{3pt}
    \begin{tabular}{|c|c|c|}
    \hline
    Parameter & Nominal value & Perturbed value  \\
    \hline
    $m$ & 0.02 & 0.015 \\
    \hline
    $I_{xx}$ & 1.395 $\times 10^{-5}$  & 2.0 $\times 10^{-5}$\\
    \hline
    $I_{yy}$ & 1.395 $\times 10^{-5}$& 1.0 $\times 10^{-5}$\\
    \hline
    $I_{zz}$ & 2.173 $\times 10^{-5}$& 3.0 $\times 10^{-5}$\\
    \hline
    $C_T$ & 3.158 $\times 10^{-10}$& 2.5 $\times 10^{-10}$\\
    \hline
    $C_D$ & 7.9379 $\times 10^{-12}$& 9.0 $\times 10^{-12}$\\
    \hline
    $d$ & 0.03973 & 0.05 \\
    \hline
    \end{tabular}
    \label{tab:param table}
\end{table}
As can be seen from Figure \ref{fig:comparison NN pert}, the prediction accuracy of the model-free FDI mechanism is unaffected by a change in the model parameters, while that of the model-based FDI mechanism drops significantly. Thus, in the scenarios when a correct system model is not known or the system dynamics undergo changes during operation, a model-based FDI mechanism might not remain reliable. Figures \ref{fig:res only vs all} and \ref{fig:comparison NN pert} illustrate that the proposed model-free NN-FDI is agnostic to the type of input used for data generation as well as to perturbations in model parameters. 

\section{Conclusion}
In this paper, we propose a learning method for effectively learning a model-free output-based FDI for the prediction of a variety of partial losses in actuation for quadrotors. The proposed NN-based FDI can verify the actuator performance with very high accuracy and correctly predicts a variety of faults. The numerical experiments demonstrated that the applicability of a model-based FDI mechanism is very limited, while that of the proposed model-free is quite broad and general. Additionally, the robustness to out-of-distribution input data illustrates that the proposed model-free mechanism can be easily trained on simple input data (e.g., LQR input), does not require the model information and generalizes to both out-of-distribution input data as well as changes in model parameters (or modeling uncertainties).

As part of future work, we will explore methods that can incorporate more general fault models where the faulty actuator can take any arbitrary signal, and more than one actuator can undergo failure simultaneously. We will also explore applications of this framework to resilient control of networked and distributed control systems, which introduce additional notions of system failure, including loss of entire nodes or communication links in addition to input disturbances and loss of control authority.

\bibliographystyle{IEEEtran}
\bibliography{refs.bib}

\begin{thebibliography}{10}
\providecommand{\url}[1]{#1}
\csname url@rmstyle\endcsname
\providecommand{\newblock}{\relax}
\providecommand{\bibinfo}[2]{#2}
\providecommand\BIBentrySTDinterwordspacing{\spaceskip=0pt\relax}
\providecommand\BIBentryALTinterwordstretchfactor{4}
\providecommand\BIBentryALTinterwordspacing{\spaceskip=\fontdimen2\font plus
\BIBentryALTinterwordstretchfactor\fontdimen3\font minus
  \fontdimen4\font\relax}
\providecommand\BIBforeignlanguage[2]{{%
\expandafter\ifx\csname l@#1\endcsname\relax
\typeout{** WARNING: IEEEtran.bst: No hyphenation pattern has been}%
\typeout{** loaded for the language `#1'. Using the pattern for}%
\typeout{** the default language instead.}%
\else
\language=\csname l@#1\endcsname
\fi
#2}}

\bibitem{lin2020human}
G.~Lin, H.~Li, H.~Ma, D.~Yao, and R.~Lu, ``Human-in-the-loop consensus control
  for nonlinear multi-agent systems with actuator faults,'' \emph{IEEE/CAA
  Journal of Automatica Sinica}, vol.~9, no.~1, pp. 111--122, 2020.

\bibitem{abbaspour2020survey}
A.~Abbaspour, S.~Mokhtari, A.~Sargolzaei, and K.~K. Yen, ``A survey on active
  fault-tolerant control systems,'' \emph{Electronics}, vol.~9, no.~9, p. 1513,
  2020.

\bibitem{eltrabyly2021fault}
A.~Eltrabyly, D.~Ichalal, and S.~Mammar, ``Fault-tolerant model predictive
  control trajectory tracking for a quadcopter with 4 faulty actuators,''
  \emph{IFAC-PapersOnLine}, vol.~54, no.~4, pp. 141--146, 2021.

\bibitem{sharifi2010fault}
F.~Sharifi, M.~Mirzaei, B.~W. Gordon, and Y.~Zhang, ``Fault tolerant control of
  a quadrotor uav using sliding mode control,'' in \emph{2010 conference on
  control and Fault-Tolerant Systems (SysTol)}.\hskip 1em plus 0.5em minus
  0.4em\relax IEEE, 2010, pp. 239--244.

\bibitem{freddi2011feedback}
A.~Freddi, A.~Lanzon, and S.~Longhi, ``A feedback linearization approach to
  fault tolerance in quadrotor vehicles,'' \emph{IFAC proceedings volumes},
  vol.~44, no.~1, pp. 5413--5418, 2011.

\bibitem{wang2020active}
B.~Wang, Y.~Shen, and Y.~Zhang, ``Active fault-tolerant control for a quadrotor
  helicopter against actuator faults and model uncertainties,'' \emph{Aerospace
  Science and Technology}, vol.~99, p. 105745, 2020.

\bibitem{zhu2021adaptive}
X.~Zhu, J.~Chen, and Z.~H. Zhu, ``Adaptive learning observer for spacecraft
  attitude control with actuator fault,'' \emph{Aerospace Science and
  Technology}, vol. 108, p. 106389, 2021.

\bibitem{smaeilzadeh2019finite}
S.~M. Smaeilzadeh and M.~Golestani, ``A finite-time adaptive robust control for
  a spacecraft attitude control considering actuator fault and saturation with
  reduced steady-state error,'' \emph{Transactions of the Institute of
  Measurement and Control}, vol.~41, no.~4, pp. 1002--1009, 2019.

\bibitem{slotine_li_1991}
J.-J.~E. Slotine and W.~Li, \emph{Applied nonlinear control: an
  introduction}.\hskip 1em plus 0.5em minus 0.4em\relax Prentice-Hall, 1991.

\bibitem{rmpc}
A.~Bemporad and M.~Morari, ``Robust model predictive control: A survey,'' in
  \emph{Robustness in identification and control}, A.~Garulli and A.~Tesi,
  Eds.\hskip 1em plus 0.5em minus 0.4em\relax Springer London, 1999, pp.
  207--226.

\bibitem{JB2022}
J.-B. Bouvier and M.~Ornik, ``Quantitative resilience of linear systems,'' in
  \emph{20th European Control Conference}, 2022, pp. 477--482.

\bibitem{er2018novel}
M.~J. Er, C.~Deng, and N.~Wang, ``A novel fuzzy logic control method for
  multi-agent systems with actuator faults,'' in \emph{2018 IEEE International
  Conference on Fuzzy Systems (FUZZ-IEEE)}.\hskip 1em plus 0.5em minus
  0.4em\relax IEEE, 2018, pp. 1--7.

\bibitem{lanzon2014flight}
A.~Lanzon, A.~Freddi, and S.~Longhi, ``Flight control of a quadrotor vehicle
  subsequent to a rotor failure,'' \emph{Journal of Guidance, Control, and
  Dynamics}, vol.~37, no.~2, pp. 580--591, 2014.

\bibitem{yang2020adaptive}
H.~Yang, Y.~Jiang, and S.~Yin, ``Adaptive fuzzy fault-tolerant control for
  markov jump systems with additive and multiplicative actuator faults,''
  \emph{IEEE Transactions on Fuzzy Systems}, vol.~29, no.~4, pp. 772--785,
  2020.

\bibitem{hwang2009survey}
I.~Hwang, S.~Kim, Y.~Kim, and C.~E. Seah, ``A survey of fault detection,
  isolation, and reconfiguration methods,'' \emph{IEEE transactions on control
  systems technology}, vol.~18, no.~3, pp. 636--653, 2009.

\bibitem{puchalski2022uav}
R.~Puchalski and W.~Giernacki, ``Uav fault detection methods,
  state-of-the-art,'' \emph{Drones}, vol.~6, no.~11, p. 330, 2022.

\bibitem{fourlas2021survey}
G.~K. Fourlas and G.~C. Karras, ``A survey on fault diagnosis methods for
  uavs,'' in \emph{2021 International Conference on Unmanned Aircraft Systems
  (ICUAS)}.\hskip 1em plus 0.5em minus 0.4em\relax IEEE, 2021, pp. 394--403.

\bibitem{guzman2019actuator}
J.~A. Guzm{\'a}n-Rabasa, F.~R. Lopez-Estrada, B.~M. Gonz{\'a}lez-Contreras,
  G.~Valencia-Palomo, M.~Chadli, and M.~Perez-Patricio, ``Actuator fault
  detection and isolation on a quadrotor unmanned aerial vehicle modeled as a
  linear parameter-varying system,'' \emph{Measurement and Control}, vol.~52,
  no. 9-10, pp. 1228--1239, 2019.

\bibitem{ouadine2020uav}
A.~Y. Ouadine, M.~Mjahed, H.~Ayad, and A.~El~Kari, ``Uav quadrotor fault
  detection and isolation using artificial neural network and
  hammerstein-wiener model,'' \emph{Studies in Informatics and Control},
  vol.~29, no.~3, pp. 317--328, 2020.

\bibitem{avram2017quadrotor}
R.~C. Avram, X.~Zhang, and J.~Muse, ``Quadrotor actuator fault diagnosis and
  accommodation using nonlinear adaptive estimators,'' \emph{IEEE Transactions
  on Control Systems Technology}, vol.~25, no.~6, pp. 2219--2226, 2017.

\bibitem{zhong2018robust}
Y.~Zhong, Y.~Zhang, W.~Zhang, J.~Zuo, and H.~Zhan, ``Robust actuator fault
  detection and diagnosis for a quadrotor uav with external disturbances,''
  \emph{IEEE Access}, vol.~6, pp. 48\,169--48\,180, 2018.

\bibitem{cen2014robust}
Z.~Cen, H.~Noura, T.~B. Susilo, and Y.~A. Younes, ``Robust fault diagnosis for
  quadrotor uavs using adaptive thau observer,'' \emph{Journal of Intelligent
  \& Robotic Systems}, vol.~73, pp. 573--588, 2014.

\bibitem{ghrib2020hybrid}
Z.~Ghrib, R.~Jaziri, and R.~Romdhane, ``Hybrid approach for anomaly detection
  in time series data,'' in \emph{2020 international joint conference on neural
  networks (ijcnn)}.\hskip 1em plus 0.5em minus 0.4em\relax IEEE, 2020, pp.
  1--7.

\bibitem{provotar2019unsupervised}
O.~I. Provotar, Y.~M. Linder, and M.~M. Veres, ``Unsupervised anomaly detection
  in time series using lstm-based autoencoders,'' in \emph{2019 IEEE
  International Conference on Advanced Trends in Information Theory
  (ATIT)}.\hskip 1em plus 0.5em minus 0.4em\relax IEEE, 2019, pp. 513--517.

\bibitem{ergen2019unsupervised}
T.~Ergen and S.~S. Kozat, ``Unsupervised anomaly detection with lstm neural
  networks,'' \emph{IEEE transactions on neural networks and learning systems},
  vol.~31, no.~8, pp. 3127--3141, 2019.

\bibitem{park2021data}
J.-H. Park and D.~E. Chang, ``Data-driven fault detection and isolation of
  system with only state measurements and control inputs using neural
  networks,'' in \emph{2021 21st International Conference on Control,
  Automation and Systems (ICCAS)}.\hskip 1em plus 0.5em minus 0.4em\relax IEEE,
  2021, pp. 108--112.

\bibitem{bondyra2022acoustic}
A.~Bondyra, M.~Ko{\l}odziejczak, R.~Kulikowski, and W.~Giernacki, ``An acoustic
  fault detection and isolation system for multirotor uav,'' \emph{Energies},
  vol.~15, no.~11, p. 3955, 2022.

\bibitem{alippi2016model}
C.~Alippi, S.~Ntalampiras, and M.~Roveri, ``Model-free fault detection and
  isolation in large-scale cyber-physical systems,'' \emph{IEEE Transactions on
  Emerging Topics in Computational Intelligence}, vol.~1, no.~1, pp. 61--71,
  2016.

\bibitem{bakhtiaridoust2022model}
M.~Bakhtiaridoust, M.~Yadegar, N.~Meskin, and M.~Noorizadeh, ``Model-free
  geometric fault detection and isolation for nonlinear systems using koopman
  operator,'' \emph{IEEE Access}, vol.~10, pp. 14\,835--14\,845, 2022.

\bibitem{thirumarimurugan2016comparison}
M.~Thirumarimurugan, N.~Bagyalakshmi, and P.~Paarkavi, ``Comparison of fault
  detection and isolation methods: A review,'' in \emph{2016 10th International
  Conference on Intelligent Systems and Control (ISCO)}.\hskip 1em plus 0.5em
  minus 0.4em\relax IEEE, 2016, pp. 1--6.

\bibitem{budaciu2019evaluation}
C.~Budaciu, N.~Botezatu, M.~Kloetzer, and A.~Burlacu, ``On the evaluation of
  the crazyflie modular quadcopter system,'' in \emph{2019 24th IEEE
  International Conference on Emerging Technologies and Factory Automation
  (ETFA)}.\hskip 1em plus 0.5em minus 0.4em\relax IEEE, 2019, pp. 1189--1195.

\bibitem{Ames2014}
A.~D. Ames, J.~W. Grizzle, and P.~Tabuada, ``Control barrier function based
  quadratic programs with application to adaptive cruise control,'' in
  \emph{53th IEEE Conference on Decision and Control}, 2014, pp. 6271 -- 6278.

\end{thebibliography}

\end{document}